\pdfoutput=1

\documentclass[11pt]{article}

\usepackage{emnlp2021}
\definecolor{ForestGreen}{RGB}{34,139,34}

\usepackage{times}
\usepackage{latexsym}
\usepackage{amsmath}
\usepackage{amssymb}
\usepackage{multirow}
\usepackage{booktabs}
\usepackage{float}
\usepackage[T1]{fontenc}

\usepackage[utf8]{inputenc}
\usepackage{multirow}

\usepackage{microtype}
\usepackage{graphicx}
\usepackage{array}
\usepackage{enumitem}
\usepackage{mathtools}
\DeclarePairedDelimiter\abs{\lvert}{\rvert}%
\DeclarePairedDelimiter\norm{\lVert}{\rVert}%

\makeatletter
\let\oldabs\abs
\def\abs{\@ifstar{\oldabs}{\oldabs*}}
\let\oldnorm\norm
\def\norm{\@ifstar{\oldnorm}{\oldnorm*}}
\makeatother

\title{Enriching and Controlling Global Semantics for Text Summarization}

\author{Thong Nguyen$^{1, 3}$\thanks{~~Work done during internship at VinAI Research}~~, Anh Tuan Luu$^{2}$\thanks{~~Corresponding Author}~~, Truc Lu$^{3}$, Tho Quan$^{3}$ \\
  $^1$VinAI Research, Vietnam \\
  $^2$Nanyang Technological University, Singapore \\
  $^3$Ho Chi Minh City University of Technology, Vietnam National University Ho Chi Minh City \\
  \texttt{\small v.thongnt66@vinai.io, tuancs219@yahoo.com, qttho@hcmut.edu.vn} \\}
  
\begin{document}

\maketitle
\begin{abstract}
Recently, Transformer-based models have been proven effective in the abstractive summarization task by creating fluent and informative summaries. Nevertheless, these models still suffer from the short-range dependency problem, causing them to produce summaries that miss the key points of document. In this paper, we attempt to address this issue by introducing a neural topic model empowered with normalizing flow to capture the global semantics of the document, which are then integrated into the summarization model. In addition, to avoid the overwhelming effect of global semantics on contextualized representation, we introduce a mechanism to control the amount of global semantics supplied to the text generation module. Our method outperforms state-of-the-art summarization models on five common text summarization datasets, namely CNN/DailyMail, XSum, Reddit TIFU, arXiv, and PubMed. 

\end{abstract}

\section{Introduction}

{\renewcommand{\arraystretch}{1.1}
\begin{table}[h!]
\centering
\begin{small}
\begin{tabular}{|p{7.5cm}|}
\hline\textbf{DOCUMENT}: While Richie Benaud rose from the suburbs to captain Australia, he will be remembered forever for his mastery of commentating. \textcolor{orange}{The champion leg spinner turned cricket commentating into an art form, earning him the title of 'the Voice of Cricket.'} \textcolor{blue}{His \textbf{commentary} was understated, measured and often extremely funny, and were perfectly timed}. Scroll down for video. \textcolor{red}{84-year-old cricket commentator Richie Benaud has passed away after a battle with skin cancer} . His sayings from the hundreds of Test and One Day cricket matches he commentated on across the world were often what fans remembered from important moments. \textcolor{purple}{His signature one liners soon dropped to a simple word. 'Marvellous...' will forever be linked to the cricket legend}. On commentating, Benaud said: 'My mantra is - put your brain into gear and if you can add to what's on the screen then do it, otherwise shut up.' He once described the scene on the field: 'From our broadcast box you can’t see any grass at all, it is simply a carpet of humanity.’ On captaincy, and he was one of the best Test captains Australia ever had, Benaud was modest: 'The hallmark of a great captain is the ability to win the toss, at the right time.’ The former leg spinner turned cricket commentating into an art form, giving him the title 'the Voice of Cricket'. But he cautioned that description with: 'Captaincy is 90 per cent luck and 10 per cent skill. But don’t try it without that 10 per cent.’ [...] \\ \hline
\textbf{GOLD SUMMARY}: \textcolor{red}{Cricket commentator Richie Benaud has passed away after cancer battle} . \textcolor{blue}{The 84-year-old will be remembered for his mastery of commentating} . \textcolor{orange}{The former leg spinner earned himself the title of the 'Voice of Cricket'}. \textcolor{purple}{His trademark line was 'Marvellous'.} \\ \hline
\textbf{PEGASUS}: \textcolor{orange}{The champion leg spinner turned cricket commentating into an art form, earning him the title of 'the Voice of Cricket'.} \textcolor{blue}{His commentary was understated, measured and often extremely funny, and were perfectly timed.} \\ \hline
\textbf{Our model}: \textcolor{red}{84-year-old cricket commentator Richie Benaud has passed away after a battle with skin cancer}. \textcolor{orange}{The champion leg spinner earned the title of 'the Voice of Cricket'}. \textcolor{blue}{His commentary was understated, measured and often extremely funny}. \textcolor{purple}{His trademark word, 'Marvellous...' will forever be linked to the cricket legend.} \\ \hline

\end{tabular}
\end{small}
\caption{An example of summarization outputs.}
\label{tab:example}
\end{table}}

Automatic text summarization corresponds to text understanding and text generation processes. In general, there are two main approaches to perform this task. Extractive systems \citep{liu2019fine, narayan2020stepwise, zhang2019hibert, jia2020neural} highlight salient words or sentences from the source text and form the final summary by concatenating them. On the other hand, abstractive methods \citep{see2017get, zhang2020pegasus, zou2020pre} switch among generating new words, choosing phrases from the source document, and rephrasing them. Abstractive summarization, which is the focus of this paper, is usually more advanced and closer to human-like interpretation.

\begin{figure*}[h]
  \includegraphics[width=0.95\textwidth]{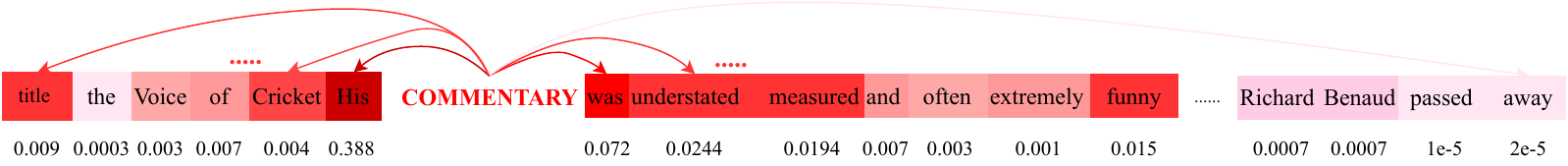} 
  \vspace{-2mm}
  \caption{Self-attention weights of “\emph{commentary}” in the PEGASUS model”}
  \label{fig:self_attention}
\end{figure*}

Recently, abstractive summarization studies \citep{lewis2019bart, zhang2020pegasus, chen2020multi} are dominated by Transformer-based architecture \citep{vaswani2017attention}. Despite good performance in large scale datasets, Transformer-based summarization models have been proven to have the tendency to favor encoding short-range dependencies \citep{zhang2020pegasus}, i.e., whenever there is one word from the input generated in the summary, the  model tends to continue generating the nearby words due to their high attention scores to the previous generated word.  As such, if the main content of the document is out of reach from the generated word, the final summary can miss that key information. For example, in Table \ref{tab:example}, PEGASUS, a state-of-the-art Transformed-based model, failed to capture one key information of the document, i.e., ``\textit{84-year-old cricket commentator Richie Benaud has passed away after a battle with skin cancer}''. To understand this phenomenon, we visualize the attention scores in the model during the generation process. As shown in Figure \ref{fig:self_attention}, when the model generates ``\textit{commentary}'', the main subject of the blue sentence, it tends to point to and generate nearby words such as ``\emph{his}'', ``\emph{understated}'', ``\emph{funny}'', etc. due to their high attention scores while words in the further range such as ``\emph{Richard}'', ``\emph{Benaud}'', ``\emph{pass}'', and ``\emph{away}'' receive little weight. Consequently, although PEGASUS generates a grammatically correct summary, the summary lacks the key content which describes the death of ``\emph{Richie Bernaud}''.

To avoid missing key points in summarizing, one solution is to furnish the models with global semantics by using probabilistic topic models such as LDA \citep{narayan2018don}, Poisson factor analysis \citep{wang2020friendly}, or  inner hidden states \citep{liu2019topic}. Nevertheless, traditional topic models were shown to be inefficient in scalability for large-scale datasets \citep{hoffman2013stochastic, rezende2015variational} and have limited capability of describing documents \citep{ding2018coherence}. 

To overcome the above problems, we propose a novel method that integrates neural topic model into summarization architecture. Specifically, we aim to utilize the posterior distribution learned from the  neural topic model as an approximation of global semantics of the document and from that, provide a signal for summarization model to have a better understanding of overall document. However, there is one critical question: how can we match the neural topic model's posterior distribution with the true posterior as it has been proven in improving the performance of variational inference \citep{rezende2015variational}? To this end, we propose a method to adapt normalizing flow in the neural topic model to have a better approximation of true distribution and integrate it into the summarization model. Integrating flow mechanism to better approximate the true posterior has been proven to improve performance for variational inference \cite{rezende2015variational} as well as for downstream tasks such as image synthesis \cite{kingma2016improved}, etc. However, to the best of our knowledge, there is no study to investigate the benefit of flow mechanism for the abstractive summarization task.


On the other hand, even though rich global semantics is beneficial, there are recent studies showing that the redundant amount of global semantics may cause harm to hidden representation since it introduces detrimental noise to the model \citep{tenney2019you, li2020incorporating}. Therefore, we propose a novel contextualized gating mechanism to control the flow of global semantics and maintain important information of the hidden states in the main summarization model.

The contributions of our paper can be summerized as follows:
\begin{itemize}[leftmargin=*]
\setlength \itemsep{-0.2em}
    \item We propose a novel architecture which takes the global semantics into consideration when performing abstractive summarization.
    \item To this end, we introduce a neural topic model which is enpowered with normalizing flow to enrich the global semantics and contextualized gating mechanism to better control the effect of global semantics on hidden representations.
    \item We conduct extensive experiments and outperform other state-of-the-art summarization models on five benchmark datasets, i.e. CNN/DailyMail, XSum, Reddit TIFU, PubMed, and arXiv, while generating summaries which favor human judgements, and producing human-interpretable topics.
\end{itemize}
\section{Related Work}
\subsection{Transformer-based Text Summarization}

Transformer \cite{vaswani2017attention} and its variants have demonstrated high efficiency in text summarization. \cite{liu2019text} first use to perform extractive summarization. \cite{zhong2020extractive} propose using Siamese BERT to score among summaries extracted from the source document, exploring the rich semantic space that those summaries are projected onto. \cite{narayan2020stepwise} combine HiBERT and structured transformers to extract the document incrementally to form the final summary. 

For abstractive approaches, \cite{zhang2020pegasus} develop a pretraining scheme well-suited for abstractive summarization. Other frameworks uniting language understanding and text generation such as BART \cite{lewis2019bart}, UniLM \cite{dong2019unified}, T5 \cite{raffel2019exploring}, \cite{tuan2020capturing}, and MASS \cite{song2019mass} provide further standards for future works. Unified system such as BottomUp \cite{gehrmann2018bottom} extracts salient phrases and then generates the summary based upon the extracted content. \cite{subramanian2019extractive} further improve with their decoder as a Transformer language model.
\subsection{Topic-aware Summarization Models}
Various works integrate global semantics of topic model into the sequential information. One method is to attend topic vectors with the hidden states, only choosing entries with high document-level representations \cite{zheng2020topic}. \cite{wang2020friendly} design three modules to incorporate topic information to attentive heads, provide topical embedding, and form document-related representations. Other works integrate topical information into convolutional-based models \cite{narayan2018don, wang2018reinforced}. \citealt{ailem2019topic} have their pointer-generator conditioned on both the input document and the latent vector. \cite{fu2020document} study how to effectively assist deep-learning summarization frameworks with external global information. Arguing that each paragraph in the document possesses a separate subtopic, they propose to merge topic information hierarchically with the dense word embedding. 

Unfortunately, there is still limited effort controlling the effect of global semantics on  the contextualized representations and enriching the global semantics for summarization performance.

\begin{figure*}[h]
    \centering
  \includegraphics[width=0.9\textwidth]{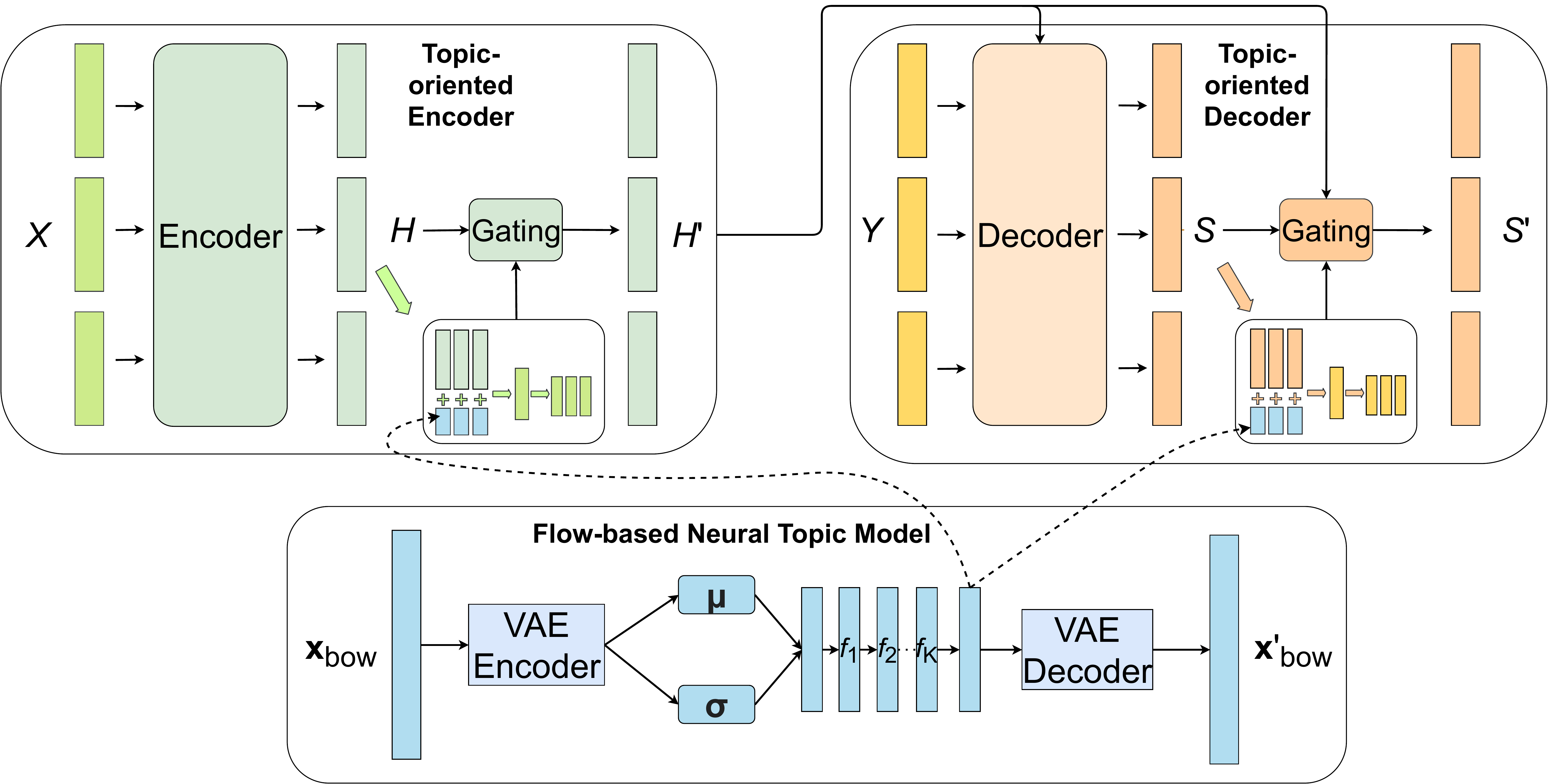} 
  \caption{Our overall architecture}
  \label{fig:model}
\end{figure*}

\section{Methodology}
The overall architecture of our approach is given in Figure \ref{fig:model}. It comprises of a topic-oriented encoder, a topic-oriented decoder, and a flow-based neural topic model. 

Formally, given a document as input, we process it into a sequence of tokens $X = \{x_i\}$, and the bag-of-word (BoW) representation $\textbf{x}_{bow}$. $X$ is taken as the input for the text summarization module, while $\textbf{x}_{bow}$ serves as the input for the neural topic model.

\subsection{Flow-based Neural Topic Model}

The architecture of the neural topic model (NTM) takes inspiration from \cite{miao2017discovering} based on variational autoencoder \cite{kingma2013auto}. In this work, we adapt the normalizing flow to the neural topic model to better grasp the global semantic patterns of the document.

\noindent \textbf{BoW Encoder.} In particular, the input $\textbf{x}_{bow}$ is first encoded into a latent variable $\mathbf{z}$ by a topic encoder. Each input is passed to obtain the prior mean $\mu$ and prior standard deviation $\sigma$
\vspace{-2mm}
\begin{equation}
    \pi = f_{MLP}(\textbf{x}_{bow}), \mu = f_1(\pi), \log \sigma = f_2(\pi)
\end{equation}
where $f_{MLP}$ is a non-linear transformation with a $\tanh$ activation function; $f_1$ and $f_2$ are two linear transformations with bias. To obtain the topic distribution, we draw the latent variable $\textbf{z} \sim \mathcal{N}(\mu, \sigma^2)$.

\noindent \textbf{Flow.} Different from conventional neural topic model, a flow is applied to map the latent vector to a more complicated distribution. Formally, flow is a chain of transformations $f_1, f_2, …, f_K$ which are all invertible and have the Jacobian easy to compute.
\vspace{-4mm}
\begin{equation}
    \textbf{z}_K = f_0 \circ f_1 ... \circ f_K (\mathbf{z}) 
\end{equation}

\noindent \textbf{BoW Decoder}. Given the new topic vector, the BoW decoder retains the original input $\textbf{x}_{bow}$ by generating $\textbf{x}'_{bow}$.  We take the following procedure to simulate the reconstruction of $\textbf{x}_{bow}$

\begin{itemize}
	\item Topic mixture $\theta = \text{softmax}(f_{\theta}(\textbf{z}_K))$
	\item For each word $w \in \textbf{x}_{bow}$, draw $w \sim \text{softmax}(f_{\phi}(\theta))$
\end{itemize}

where $f_*(\cdot)$ is a ReLU-activated non-linear transformation. The weight matrix of $f_{\phi}$ is chosen as the topic-word distribution $(\phi_1, \phi_2, ..., \phi_K)$. We proceed to employ the topic mixture $\theta$ to guide the text summarization process.

\subsection{Neural Topic Modeling for Transformer}

Text summarization model is passed a source document $X = \{x_i\}_{i=1}^{N}$ and its task is to predict the target summary $Y = \{y_j\}_{j=1}^{M}$. In this setting, the document $D$ has $N$ tokens and the summary $S$ has $M$ tokens ($M < N$).

Our model inherits the Transformer-based architecture. Particularly, it consists of an encoder and decoder. The encoder learns the context of the source text, and the decoder then predicts the target summary, by learning the context of the generated tokens and attending over encoder hidden states. In our case, we make both the encoder and decoder conditioned on the latent topic yielded by the neural topic model.

\noindent \textbf{Topic-oriented Encoder} We add the special token “\emph{CLS}” to the beginning of the input. At each iteration, the encoder outputs a localized representation $H = \{\textbf{h}_i\}_{i=1}^{N}$ for each token in the source document $X$
\vspace{-2mm}
\begin{equation}
    \textbf{h}_{CLS}, \textbf{h}_1, ..., \textbf{h}_N  = \text{Encoder}(x_{CLS}, x_1, ..., x_N)
\end{equation}

This explores the relationship among the input tokens (the encoder), or discovering the context each token stays in. We relate the context of each word to the main topic of the document by modulating the $i$-th hidden state $\textbf{h}_i$
\vspace{-2mm}
\begin{equation}
    \textbf{h}'_i = g(\textbf{h}_i, \theta)
\end{equation}
where $g$ is a function used to introduce the global semantics to the hidden representations which we will discuss later as the contextualized gating mechanism in section \ref{sec:contextualized_gating}

\noindent \textbf{Topic-oriented Decoder} We also make “\emph{CLS}” the first input of the decoder. The decoder bridges the summary $Y$ and document $X$, creating target hidden states $S = \{\textbf{s}_j\}_{j=1}^{M}$ aligned with the source text. Because of the uni-directionality of the text summarization task, the decoder must work in a left-to-right fashion
\begin{equation}
    \textbf{s}_j  = \text{Decoder}(y_{CLS}, y_1,y_2, ..., y_{j-1}, \{\textbf{h}'_i\}_{i=1}^{N}) 
\end{equation}
Similar to the Encoder, we seek to inject the semantics of the topic model into the output hidden state.
\begin{equation}
    \textbf{s}'_{j} = g(\{\textbf{h}'_i\}_{i=1}^{N}, \textbf{s}_{j}, \theta)
\end{equation}

\subsection{Contextualized Gating Mechanism}
\label{sec:contextualized_gating}
Because a specified amount of semantic meaning, whether it is local or global, has been embedded in the contextualized representations, it is reasonable to only append sufficient information to the calculated hidden states to maximize the efficiency of the topical information. We adapt the gating mechanism of \cite{cho2014properties} to achieve this goal. In our contextualized gating mechanism, we approximate the necessary amount of global semantics based on the obtained hidden states.

\noindent \textbf{Encoder Gating} For the encoder, we take the hidden representation of “\emph{CLS}” token to control the amount of additional global information
\begin{equation}
    \lambda^E = \text{Sigmoid}(W^E \textbf{h}_{CLS} + b^E) 
\end{equation}
where $W^E \in \mathbb{R}^{d \times d}$, and $d$ is the dimension of the hidden representation. We form the topic-aware hidden state by merging it with the topic mixture and mapping it onto a topical space
\begin{gather}
    \textbf{u}_i = [\textbf{h}_i, \theta] \\
    \textbf{c}_i = f_{enc\_topic}(\textbf{u}_i) 
\end{gather}

where $f_{enc\_topic}$ is a non-linear transformation. The topic-oriented encoder hidden state of every token is the fusion of the topic-aware and the original representation.
\begin{equation}
    \textbf{h}'_i = \lambda^E \textbf{c}_i + (1-\lambda^E) \textbf{h}_i
\end{equation}

\noindent \textbf{Decoder Gating} The amount of topic mixture used for the decoder is controlled by both encoder and decoder hidden state
\begin{equation}
    \lambda^D = \text{Sigmoid}(W_1^D \textbf{h}_{CLS} + W_2^D \textbf{s}_{CLS} + b^D) \\
\end{equation}
where $W_1^D \in \mathbb{R}^{d \times d}$,  $W_2^D \in \mathbb{R}^{d \times d}$. This switching probability is used to modulate the decoder hidden state, which follows the same computation with the encoder gating.
\begin{gather}
    \textbf{t}_j = [\textbf{s}_j, \theta_{dec}] \\
    \textbf{e}_j = f_{dec\_topic}(\textbf{t}_j) \\
    \textbf{s}'_j = \lambda^D \textbf{e}_j + (1-\lambda^D) \textbf{s}_j
\end{gather}

\subsection{Training Objective}
Our framework favors end-to-end learning of neural topic modeling and text summarization. In this section, we formally define the objective functions for the two modules.

For our neural topic model, the objective function is derived from the evidence lower bound \cite{blei2017variational}. We adapt the change of variables in normalizing flow that determine the distribution of the variable at the end of the flow to the loss of neural topic model
\vspace{-3mm}
\begin{equation}
    \begin{split}
        &\mathcal{L}_{\text{NTM}} \\
        &= \log p(\mathbf{x,z}) - \log q(\mathbf{z}|\mathbf{x}) \\
        &= -\log q(\mathbf{z}_0) + \sum_{i=1}^{K} \log \abs{\det \frac{\partial f_i}{\partial z_{i-1}}} \\ 
        & +\log p(\mathbf{x}|\mathbf{z}_K) + \log p(\mathbf{z}_K)
    \end{split}
\end{equation}
where $p(\mathbf{z}_K)$ denotes the prior distribution constructed by the flow; $p(\mathbf{x}|\mathbf{z}_K)$ stands for the log-likelihood of the document;  log $q_K(\mathbf{z}_K)$ denotes the approximate posterior distribution. Detailed derivation is available in Appendix.

For text summarization, we minimize the cross-entropy loss 
\begin{equation}
    \mathcal{L}_{\text{sum}} = -\sum_{j=1}^{M} \log p(y_j|\{x_i\}_{i=1}^{N}, y_{i<j})
\end{equation}

where $N$ and $M$ are the length of the document $X$ and summary $Y$, respectively.
The entire framework is trained with the linear combination of two loss functions $\mathcal{L}_{\text{sum}}$ and $\mathcal{L}_{\text{NTM}}$
\begin{equation}
    \mathcal{L} = \mathcal{L}_{\text{sum}} + \lambda \mathcal{L}_{\text{NTM}}
    \label{eq:loss}
\end{equation}
where $\lambda$ is the hyperparameter balancing the effect of neural topic model on the training process.

\section{Experimental Setup}

\subsection{Datasets}

We evaluate our proposed method on five benchmark datasets: CNN/DailyMail (CNN/DM) \cite{hermann2015teaching}, XSum \cite{narayan2018don}, Reddit TIFU \cite{kim2018abstractive}, arXiv, and PubMed \cite{cohan2018discourse}. The datasets possess various styles and varying lengths.\\ 

\noindent \textbf{CNN/DM} is constructed by collecting news articles written by CNN and DailyMail journalists. For each article, its highlights are chosen as the summary. We use the non-anonymized version and follow the conventional training/validation/test split in \cite{hermann2015teaching}. \\

\noindent \textbf{XSum} comprises of 226,711 news articles, each of which is linked with a one-sentence summary. Our preprocessing and training/validation/test split is analogous to \cite{narayan2018don}. \\

\noindent \textbf{Reddit TIFU} includes 120K informal posts from the online discussion forum Reddit, strictly following the rule of constructing an expressive “TL;DR” summary. In this work, the long subset of the dataset is applied for performance evaluation. \\

\noindent \textbf{arXiv, PubMed} are two long-document datasets of scientific publications. For each document, the abstract is chosen to be the summary.  \\

\noindent  We present the statistics of datasets in Table \ref{tab:datasets}.
{\renewcommand{\arraystretch}{1.05}
\begin{table}[h]
\centering
\begin{small}
\begin{tabular}{c|c|c|c|c|c}
\hline
\textbf{Dataset} & \textbf{Train} & \textbf{Val} & \textbf{Test} & $l_D$ & $l_S$ \\ \hline
CNN/DM & 287113 & 13368 & 11490 & 781 & 56 \\
XSum & 204045 & 11332 & 11334 & 431 & 23 \\
Reddit TIFU & 33711 & 4214 & 4214 & 433 & 23 \\ 
arXiv & 203037 & 6436 & 6440 & 4938 & 220 \\ 
PubMed & 119924 & 6633 & 6658 & 3016 & 203 \\ 
\hline
\end{tabular}
\end{small}
\caption{Description of the evaluation datasets. $l_D$ and $l_S$ stand for average length of document and summary}
\label{tab:datasets}
\end{table}}

\subsection{Implementation Details}

\textbf{Neural Topic Model} Provided a dataset, firstly we pretrain the flow-based topic model so that it is able to obtain the prior context of the downstream documents. We experimented with different choices of the topic number $T \in \{50, 100, 200\}$ and the number of invertible transformations applied in the flow of neural topic model $K \in {1, 4, 16}$ on CNN/DailyMail dataset. 

\noindent \textbf{Summarization Module} We use the pretrained checkpoint open-sourced by \cite{zhang2020pegasus}, integrate and jointly finetune with the flow-based neural topic model on downstream datasets. Following \cite{zhang2020pegasus}, during test time, our beam search is conducted with a beam size of 8, and top-3 checkpoints are selected based on their evaluation loss in the validation set, and we average their results on the test set. More detailed settings can be found in Appendix.

\subsection{Comparisons}

As baselines, we compare our proposed architecture against a wide variety of previous studies:
\vspace{-1mm}
\begin{itemize}[leftmargin=*]
\setlength \itemsep{-0.2em}
    \item \textbf{PTGEN} \cite{see2017get}: a pointer-generator baseline that allows switching between generating words from the vocabulary and copying words from the source.
    
    \item \textbf{PTGEN+Cov} \cite{see2017get}: a pointer-generator baseline with coverage mechanism.
    
    \item \textbf{DRM} \cite{paulus2017deep}: a deep reinforced model which handles the coverage mechanism by using intra-attention among decoder tokens.
    
    \item \textbf{DiscAttn} \cite{cohan2018discourse}: a Seq2Seq model which targets the long-document summarization.
    
    \item \textbf{BertSUM} \cite{liu2019text}: a baseline with finetuning strategy is designed based on the discrepancy between the encoder and decoder.
    
    \item \textbf{ExtSum-LG+RdLoss} \cite{xiao2020systematically}: a Transformer-based  model with a training scheme to explicitly reduce redundancy.
    
    \item \textbf{MatchSUM} \cite{zhong2020extractive}: a baseline that makes use of Siamese BERT to score among candidate summaries.
    
    \item \textbf{BottomUp} \cite{gehrmann2018bottom}: a baseline uses extractive-abstractive approach: initially extracts salient phrases and performs abstractive summarization based on extracted content.
    
    \item \textbf{TLM-I+E} \cite{subramanian2019extractive}: a baseline improved on  \cite{gehrmann2018bottom} by utilizing Transformer language model as the decoder.
    
    \item \textbf{BART} \cite{lewis2019bart}: a baseline which is pretrained with denoising tasks.
    
    \item \textbf{PEGASUS} \cite{zhang2020pegasus}: a Transformer-based model with pretraining procedure comprised of two tasks: masked sentences prediction and masked language modeling.
    
    \item \textbf{BertSUM + TA} \cite{wang2020friendly}: a BertSUM model equipped with the topic assistant inspired by the Poisson Factor Analysis topic model.
    
    \item \textbf{BART + TA} \cite{wang2020friendly}: a BART version with a plugged-in topic assistant inspired by the Poisson Factor Analysis topic model.
    
    \item \textbf{VHTM} \cite{fu2020document}: a baseline which takes hierarchical structure of the source text into account and considers each section as a subtopic.
\end{itemize} 

\section{Experimental Results}


{\renewcommand{\arraystretch}{1.05}
\begin{table}[t]
\centering
\begin{small}
\begin{tabular}{c|c|c|c}
\hline
\textbf{Model} & \textbf{R1} & \textbf{R2} & \textbf{RL} \\ \hline
PTGEN & 36.44 & 15.66 & 33.42 \\
PTGEN + Cov & 39.56 & 17.28 & 36.38 \\
DRM & 41.16 & 15.75 & 39.08 \\ \hline
BertSUM & 43.85 & 20.34 & 39.90 \\
MatchSUM & 44.41 & 20.86 & 40.55 \\
BottomUp & 41.22 & 18.68 & 38.34 \\ 
BART & 44.16 & 21.28 & 40.90 \\
PEGASUS & 44.17 & 21.47 & 41.11 \\ \hline
VHTM & 40.57 & 18.05 & 37.18 \\
BertSUM + TA & 43.06 & 20.58 & 39.67 \\
BART + TA & 44.47 & 21.39 & 41.32 \\ \hline
Our Model & \textbf{44.52} & \textbf{21.95} & \textbf{41.39} \\
\hline
\end{tabular}
\end{small}
\caption{Results in text summarization on CNN/DailyMail}
\label{tab:text_summ_cnn_dailymail}
\end{table}}

{\renewcommand{\arraystretch}{1.05}
\begin{table}[t]
\centering
\begin{small}
\begin{tabular}{c|c|c|c}
\hline
\textbf{Model} & \textbf{R1} & \textbf{R2} & \textbf{RL} \\ \hline
PTGEN & 29.70 & 9.21 & 23.24 \\
PTGEN + Cov & 28.10 & 8.02 & 21.72 \\ \hline
BertSUM & 38.81 & 16.50 & 31.27 \\
MatchSUM & 24.86 & 4.66 & 18.41 \\
BART & 45.14 & 22.27 & 37.25 \\ 
PEGASUS & 47.21 & 24.56 & 39.25 \\ \hline
BertSUM + TA & 39.77 & 17.39 & 32.39 \\
BART + TA & 45.76 & 22.68 & 38.03 \\ \hline 
Our Model & \textbf{49.57} & \textbf{25.08} & \textbf{41.81} \\
\hline
\end{tabular}
\end{small}
\caption{Results in text summarization on XSum}
\label{tab:text_summ_xsum}
\end{table}}

{\renewcommand{\arraystretch}{1.05}
\begin{table}[h]
\centering
\begin{small}
\begin{tabular}{c|c|c|c}
\hline
\textbf{Model} & \textbf{R1} & \textbf{R2} & \textbf{RL} \\ \hline
BART & 24.19 & 8.12 & 21.31 \\
MatchSUM & 25.09 & 6.17 & 20.13 \\ 
PEGASUS & 26.63 & 9.01 & 21.60 \\ \hline
Our Model & \textbf{27.96} & \textbf{9.43} & \textbf{23.08} \\
\hline
\end{tabular}
\end{small}
\caption{Results in text summarization on Reddit TIFU}
\label{tab:text_summ_reddit_tifu}
\end{table}}

{\renewcommand{\arraystretch}{1.05}
\begin{table}[h]
\centering
\begin{small}
\begin{tabular}{c|c|c|c}
\hline
\textbf{Model} & \textbf{R1} & \textbf{R2} & \textbf{RL} \\ \hline
PTGEN + Cov & 32.06 & 9.04 & 25.16 \\
DiscAttn & 35.80 & 11.05 & 31.80 \\ \hline
ExtSum-LG+RdLoss & 44.01 & 17.79 & 39.09 \\
TLM-I+E & 41.62 & 14.69 & 38.03 \\ 
PEGASUS & 43.82 & 16.74 & 39.15 \\ \hline
Our Model & \textbf{44.53} & \textbf{19.22} & \textbf{40.61} \\
\hline
\end{tabular}
\end{small}
\caption{Results in text summarization on arXiv dataset}
\label{tab:text_summ_arXiv}
\end{table}}

{\renewcommand{\arraystretch}{1.05}
\begin{table}[h]
\centering
\begin{small}
\begin{tabular}{c|c|c|c}
\hline
\textbf{Model} & \textbf{R1} & \textbf{R2} & \textbf{RL} \\ \hline
PTGEN + Cov & 31.55 & 8.52 & 27.38 \\
DiscAttn & 38.93 & 15.37 & 35.21 \\ \hline
MatchSUM & 41.21 & 14.91 & 20.13 \\ 
ExtSum-LG+RdLoss & 45.30 & 20.42 & 40.95 \\
Sent-CLF & 42.13 & 16.27 & 39.21 \\
PEGASUS & 44.29 & 19.19 & 40.42 \\ \hline
Our Model & \textbf{45.99} & \textbf{20.49} & \textbf{41.25} \\
\hline
\end{tabular}
\end{small}
\caption{Results in text summarization on PubMed}
\label{tab:text_summ_pubmed}
\end{table}}

\subsection{Automatic Evaluation} \label{subsec:text_summ_result}

We use the automatic metrics of ROUGE scores \cite{lin2004rouge}. In Table \ref{tab:text_summ_cnn_dailymail}, \ref{tab:text_summ_xsum}, \ref{tab:text_summ_reddit_tifu}, \ref{tab:text_summ_arXiv}, and \ref{tab:text_summ_pubmed}, we report the unigram overlap (ROUGE-1), bigram overlap (ROUGE-2) to assess informativeness, and longest common subsequence (ROUGE-L) for the fluency of the generated summary. Our model outperforms prior works on five standard datasets.

For CNN/DailyMail, we achieve an absolute improvement of 0.35 in ROUGE-1, 0.48 in ROUGE-2, and 0.28 in ROUGE-L over PEGASUS. Furthermore, our model obtains better results than the previous topic-aware model BART + TA in ROUGE-2 with 0.6 points. This shows that our methods can generate summaries that include important content in the document. 

On the XSum dataset, which is more abstractive than CNN/DailyMail \cite{bommasani2020intrinsic}, our gain is even more pronounced. Compared with BART + TA, we achieve 3.8 absolute improvement in ROUGE-1, 2.4 in ROUGE-2, and 3.8 in ROUGE-L.

For Reddit TIFU, in which most of the source texts and the target summaries are informal, our model outperforms PEGASUS by 1.3 in ROUGE-1, 0.4 in ROUGE-2, and 1.5 in ROUGE-L. These results show that global semantics is capable of helping the model generate better target summaries.

For arXiv and PubMed dataset, we also achieve improvement over the baseline PEGASUS, which is designed specifically for abstractive text summarization. In particular, for arXiv dataset, we gain an increase of 0.71 in ROUGE-1, 2.48 in ROUGE-2, and 1.46 in ROUGE-L. For PubMed dataset, the increase is 1.7 in ROUGE-1, 1.3 in ROUGE-2, and 0.83 in ROUGE-L.  

\subsection{Human Evaluation}

{\renewcommand{\arraystretch}{1.05}
\begin{table}[h]
\centering
\begin{small}
\begin{tabular}{c|c|c}
\hline
\textbf{Model} & \textbf{Preference Score} & \textbf{QA score  } \\ \hline
BART & -0.286 & 24.59 \\
PEGASUS & -0.257 & 26.53 \\
Our Model & 0.250 & 46.94 \\ \hline
Gold Summary & \textbf{0.536} & \textbf{93.88} \\
\hline
\end{tabular}
\end{small}
\caption{Human evaluation}
\label{tab:human_eval}
\end{table}}

Since the automatic metric does not fully reveal the true quality of the model, we conduct a human evaluation for further assessment. To achieve that goal, we design two tests in order to elicit human judgements in two ways.

In the first experiment,  we presented summaries of PEGASUS \cite{zhang2020pegasus}, BART \cite{lewis2019bart}, our model, and the gold summary, then asked four professional English speakers to rate the summaries from worst to best in terms of informativeness, faithfulness, topic coherence, and fluency. We randomly sampled 100 summaries from 100 documents of CNN/DailyMail test set. The score of a system is equal to the percentage of times it was selected as the best minus the percentage of times it was chosen as the worst.

In the second experiment, we applied the question answering (QA) paradigm. For each document, we create two independent questions which emphasizes the key information of the text. Participants would read and answer those questions as best as they could. The score for one system is the percentage of questions the participants answer correctly. 

Ten professional English speakers were asked to participate in two assessments. The results in table \ref{tab:human_eval} show that our generated summaries favor human judgements, and are more likely to maintain the important content in the original text than other systems' summaries. 

The Fleiss’ Kappa scores with overall agreement percentages of the first and second human evaluation experiments were denoted in Table \ref{tab:human_eval}. As shown in the Table, the measures demonstrate a good inter-agreement among the annotators.

{\renewcommand{\arraystretch}{1.05}
\begin{table}[H]
\centering
\begin{small}
\begin{tabular}{c|c|c}
\hline
\textbf{Test} & \textbf{Fleiss' Kappa} & \textbf{Overall Agreement} \\ \hline
Preference & 0.61 & 70.45\% \\
QA & 0.77 & 81.13\% \\ 
\hline
\end{tabular}
\end{small}
\caption{Fleiss' Kappa and Overall Agreement percentage of each human evaluation test. Higher score indicates better agreement.}
\label{tab:human_eval}
\end{table}}

\subsection{Flow-based neural topic model with other Transformer-based model}

{\renewcommand{\arraystretch}{1.05}
\begin{table}[H]
\centering
\begin{small}
\begin{tabular}{p{3.5cm}|c|c|c}
\hline
\textbf{Model} & \textbf{R1} & \textbf{R2} & \textbf{RL} \\ \hline
BART & 44.16 & 21.28 & 40.90 \\
BART + Flow-based NTM + Gating & \textbf{44.89}  & \textbf{21.74}  & \textbf{41.48}  \\ \hline
\end{tabular}
\end{small}
\caption{Results when applying flow-based neural topic model and contextualized gating for BART on CNN/DailyMail dataset \cite{lewis2019bart}}
\label{tab:plug_in_cnn}
\end{table}}

{\renewcommand{\arraystretch}{1.05}
\begin{table}[H]
\centering
\begin{small}
\begin{tabular}{p{3.5cm}|c|c|c}
\hline
\textbf{Model} & \textbf{R1} & \textbf{R2} & \textbf{RL} \\ \hline
BART & 45.14 & 22.27 & 37.25 \\
BART + Flow-based NTM + Gating & \textbf{46.86}  & \textbf{23.74}  & \textbf{38.49}  \\ \hline
\end{tabular}
\end{small}
\caption{Results when applying flow-based neural topic model and contextualized gating for BART on XSum dataset \cite{lewis2019bart}}
\label{tab:plug_in_xsum}
\end{table}}

{\renewcommand{\arraystretch}{1.05}
\begin{table}[H]
\centering
\begin{small}
\begin{tabular}{p{3.5cm}|c|c|c}
\hline
\textbf{Model} & \textbf{R1} & \textbf{R2} & \textbf{RL} \\ \hline
BART & 43.92 & 16.36 & 39.16 \\
BART + Flow-based NTM + Gating & \textbf{47.78}  & \textbf{18.28}  & \textbf{41.47}  \\ \hline
\end{tabular}
\end{small}
\caption{Results when applying flow-based neural topic model and contextualized gating for BART on arXiv dataset \cite{lewis2019bart}}
\label{tab:plug_in_arxiv}
\end{table}}

To study the effect of our topic-oriented module on other abstractive Transformer-based model, we integrate our flow-based neural topic model and contextualized gating into BART \cite{lewis2019bart}. In particular, we continue to finetune on CNN/DailyMail, XSum, and arXiv dataset, given the pretrained checkpoint. As can be seen in Table \ref{tab:plug_in_cnn}, \ref{tab:plug_in_xsum}, \ref{tab:plug_in_arxiv}, our topic-oriented module is able to improve the performance, showing general effectiveness on other Transformer-based architecture.

\subsection{Analysis on Neural Topic Model and Traditional Topic Model}
To substantiate our hypothesis that neural topic model does enhance the summarization performance in large-scale datasets, we have conducted experiments to combine the Transformer-based summarization module with traditional topic model, i.e. Latent Dirichlet Allocation (LDA) and Poisson Factor Analysis (PFA) on CNN/DailyMail and PubMed. We denoted the results in Table \ref{tab:topic_model_cnn} and Table \ref{tab:topic_model_pubmed}. As it can be seen, neural topic models, particularly our proposed model, significantly outperform approaches of traditional topic models on abstractive summarization.

{\renewcommand{\arraystretch}{1.05}
\begin{table}[H]
\centering
\begin{small}
\begin{tabular}{p{3.5cm}|c|c|c}
\hline
\textbf{Model} & \textbf{R1} & \textbf{R2} & \textbf{RL} \\ \hline
PEGASUS + LDA + Gating & 44.17 & 21.47 & 41.11 \\
PEGASUS + PFA + Gating & 44.18  & 21.53  & 41.14 \\ 
PEGASUS + VAE + Gating & 44.33  & 21.71  & 41.27 \\ \hline
Our Model & \textbf{44.52}  & \textbf{21.95}  & \textbf{41.39} \\
\hline
\end{tabular}
\end{small}
\caption{Results when adapting various topic models on CNN/DailyMail dataset}
\label{tab:topic_model_cnn}
\end{table}}

{\renewcommand{\arraystretch}{1.05}
\begin{table}[H]
\centering
\begin{small}
\begin{tabular}{p{3.5cm}|c|c|c}
\hline
\textbf{Model} & \textbf{R1} & \textbf{R2} & \textbf{RL} \\ \hline
PEGASUS + LDA + Gating & 44.36 & 19.24 & 40.53 \\
PEGASUS + PFA + Gating & 44.41 & 19.19 & 40.55 \\ 
PEGASUS + VAE + Gating & 45.46  & 19.84  & 40.89 \\ \hline
Our Model & \textbf{45.99}  & \textbf{20.49}  & \textbf{41.25} \\
\hline
\end{tabular}
\end{small}
\caption{Results when adapting various topic models on PubMed dataset}
\label{tab:topic_model_pubmed}
\end{table}}

\subsection{Latent Topic Analysis}

{\renewcommand{\arraystretch}{1.05}
\begin{table}[H]
\centering
\begin{small}
\begin{tabular}{c|c|c}
\hline
\textbf{Datasets} & \textbf{CNN/DailyMail} & \textbf{XSum} \\ \hline
LDA & 35.03 & 26.22 \\
LSA & 41.64 & 27.69 \\
VAE-based NTM & 52.60 & 52.94  \\ \hline
Our Model & \textbf{53.25} & \textbf{53.09} \\
\hline
\end{tabular}
\end{small}
\caption{$C_V$ topic coherence score on benchmark datasets. Higher scores mean more coherent topics}
\label{tab:cv_coherence_score}
\end{table}}

It is inherent that latent vector is useful for text summarization, as shown in section \ref{subsec:text_summ_result}. Here we study whether jointly training with summarization module helps the topic model produce human-interpretable topics.

\noindent \textbf{Coherence Score Comparison} We decide to evaluate the topic models with the automatic $C_V$ measure. Following \cite{zeng2018topic, wang2020friendly}, we pick the top 10 words from each topic and average $C_V$ scores of all topics. The results are reported on two summarization datasets, CNN/DailyMail and XSum. To conduct the comparisons, we take LDA and LSA as probabilistic baselines, as they are notable and well-known for human interpretability. For both baselines, we execute 1000 iterations to assure convergence. As Table \ref{tab:cv_coherence_score} shows, our model outperforms traditional topic models, which implies that jointly training neural topic model and text summarization creates human-understandable topics. 

\noindent \textbf{Sample Topics} To further assess the quality of the topics learned by our system, we continue to extract some sample words (Table 6) indicating the context around “\emph{liverpool chelsea}” discovered by the model trained on CNN/DailyMail dataset. As can be realized, the topics pertaining to probabilistic topic models such as LSA and LDA contain some mixed topic words. Conversely, our neural topic models trained with the text summarization module produce the topic which looks more coherent. In particular, our words refer to the context which involves the teams competing in the football championship of England, such as “\emph{arsenal}”, “\emph{tottenham}”, etc. and related factors, for instance, “\emph{balls}”, “\emph{prize}”, “\emph{winning}”, etc. 

{\renewcommand{\arraystretch}{1.05}
\begin{table}[H]
\centering
\hspace{-4mm}
\begin{small}
\begin{tabular}{p{1.6cm}|p{5.7cm}}
\hline
LDA & \textcolor{red}{father} liverpool \textcolor{red}{son} chelsea called group night \textcolor{red}{child} west cup \\ 
\hline
LSA & chelsea beat half winner jose mourinho \textcolor{red}{table} \textcolor{red}{place} happy \textcolor{red}{lake} \\ \hline
VAE-based NTM & liverpool \textcolor{red}{salmon} manchester england everton newcastle bale premiership fabregas clasico \\ \hline
Our Model & liverpool cup leagues chelsea balls night tottenham prize winning arsenal \\
\hline
\end{tabular}
\end{small}
\caption{Top 10 words for the topic related to “\emph{liverpool chelsea}”. Red words highlight non-topic words.}
\label{tab:cv_coherence_score}
\end{table}}

\subsection{Ablation Study}
{\renewcommand{\arraystretch}{1.05}
\begin{table}[H]
\centering
\begin{small}
\begin{tabular}{p{4cm}|c|c|c}
\hline
\textbf{Model} & \textbf{R1} & \textbf{R2} & \textbf{RL} \\ \hline
Our Model (with Flow-based NTM and Gating) & \textbf{49.57} & \textbf{25.08} & \textbf{41.81} \\
- with VAE-based NTM and Gating & 48.13 & 23.91 & 40.68 \\
- with Flow-based NTM & 46.83 & 23.89 & 39.78 \\
- with VAE-based NTM & 46.30 & 23.59  & 39.05 \\
\hline
\end{tabular}
\end{small}
\caption{Ablation Study on XSum test set}
\label{tab:ablation}
\end{table}}

In this section, we proceed to study the impact that (1) The integration of normalizing flow and (2) The contextualized gating mechanism have on the text summarization performance.

\noindent \textbf{Impact of the contextualized gating mechanism} It can be seen that plainly incorporating the global semantics into the model makes the performance improvement drop strongly. As shown in table \ref{tab:ablation}, the ROUGE-1 score’s decreases more than 2 points compared with models we apply contextualized gating. We hypothesize that in numerous cases, the effect of global semantics overwhelm the benefits of contextualized representations.

\noindent \textbf{Impact of integrating normalizing flow} In this ablation, we eliminate the normalizing flow from the neural topic modeling. As shown in Table \ref{tab:ablation}, without the normalizing flow, the improvement that the latent vector brings is downgraded, nearly 0.4 of ROUGE-1 for using contextualized gating and 0.53 of ROUGE-1 in non-gating case . We hypothesize that the plain neural topic model does not give a sufficiently expressive global semantics as the neural topic model using normalizing flow.

\subsection{Case Studies}
Table \ref{tab:example} shows a case study on the summarization results of PEGASUS and our models. While PEGASUS model misses the key information related to the death of “\emph{Richie Benauld}”, our model successfully include it in the final summarization. It shows the effectiveness of our model in capturing key information in the document, thanks to the contribution of neural topic model and gating mechanism. Remarkably, our model is also able to rephrase “\emph{signature one liners}” as “\emph{trademark word}” when describing \emph{Richie Benauld}’s famous quote, not just copying the words in the original document. More case studies can also be found in Appendix.
\section{Conclusion}

In this paper, we propose a method to utilize global semantics for text summarization task. In particular, we aim to fit the global semantics to expressively describe the documents. Moreover, we  find that maintaining the information in the original contextualized representations is also beneficial for the summarization performance. We outperform other state-of-the-art models on five benchmark datasets.



\bibliography{anthology,emnlp2021}
\bibliographystyle{acl_natbib}
\newpage
\appendix
\onecolumn

\section{Summary examples}
\label{app:summ_examples}

We present some summary examples in this section

{\renewcommand{\arraystretch}{1.2}
\begin{table}[h!]
\centering
\begin{small}
\begin{tabular}{|p{15cm}|}
\hline
\textbf{DOCUMENT}: \textcolor{red}{New York (CNN) New York state authorities have issued a health alert following a dramatic spike in hospital visits for synthetic marijuana-related emergencies. Gov. Andrew Cuomo said Friday that more than 160 patients in nine days have been rushed to hospitals across the state for adverse reactions to synthetic cannabinoid}, known as "spice" or "K2." \textcolor{cyan}{"Spice" and other similar synthetic drugs are often marketed as legal plant material coated with chemicals that are supposed to mimic the effects of marijuana}, according to a statement from the governor's office. "Since the exact compounds contained in synthetic cannabinoid products change so frequently, it's often impossible for users to know exactly what they are putting in their body," acting New York State Health Commissioner Dr. Howard Zucker said. \textcolor{ForestGreen}{Symptoms after use have a wide range of severity, from confusion, drowsiness and headaches to increased heart rate, seizures and loss of consciousness}, according to the New York State Department of Health. \textcolor{pink}{Synthetic marijuana is popular among teens because it is marketed as incense or natural products to "mask its true purpose"}, the health department statement said. \textcolor{orange}{"Young people may be fooled into thinking that these substances are safe because they are sold over the counter or are in colorful packaging}, but they are not made for human consumption," \textcolor{blue}{New York Alcohol and Substance Abuse Service s Commissioner Arlene Gonzalez Sanchez said. "They are dangerous and can have significant, long-term effects on the brain." The recent surge is not isolated in New York; other states across the country have noticed similar trends. Alabama Department of Public Health issued a statement last week acknowledging a rise of synthetic marijuana usage and said there had been 98 overdoses suspected to be linked with "spice" in the previous month. Mobile County alone has seen seven cases in 2015 so far, more than the entire previous year, the statement said. Mississippi health officials are also concerned that synthetic marijuana is on the rise.} [...] \\ \hline
\textbf{GOLD SUMMARY}: \textcolor{red}{New York reports 160 hospitalizations related to synthetic marijuana . Gov. Andrew Cuomo issued a health alert .} \\ \hline
\textbf{PEGASUS}: \textcolor{pink}{Synthetic marijuana is popular among teens because it is marketed as incense or natural products to "mask its true purpose".} \textcolor{cyan}{"Spice" and other similar synthetic drugs are often marketed as legal plant material coated with chemicals that are supposed to mimic the effects of marijuana.} \textcolor{ForestGreen}{Symptoms after use have a wide range of severity, from confusion, drowsiness and headaches to increased heart rate, seizures and loss of consciousness.} \\ \hline
\textbf{BART}: \textcolor{cyan}{"Spice" and other similar synthetic drugs are often marketed as legal plant material.} \textcolor{ForestGreen}{Symptoms after use have a wide range of severity, from confusion, drowsiness and headaches to increased heart rate, seizures and loss of consciousness.} \textcolor{orange}{"Young people may be fooled into thinking that these substances are safe because they are sold over the counter or are in colorful packaging," official says.} \\ \hline
\textbf{Our model}: \textcolor{red}{New York Gov. Andrew Cuomo says 160 people have been hospitalized in nine days}. \textcolor{ForestGreen}{Symptoms after use include confusion, drowsiness, headaches, increased heart rate, seizures and loss of consciousness}. \textcolor{blue}{Health officials are concerned that synthetic marijuana is on the rise.} \\ \hline

\end{tabular}
\end{small}
\caption{A generated summary example .}
\label{tab:example_1}
\end{table}}

{\renewcommand{\arraystretch}{1.2}
\begin{table}[h!]
\centering
\begin{small}
\begin{tabular}{|p{15cm}|}
\hline
\textbf{DOCUMENT}: (CNN)Panic. Tears. Fear. All those feelings and more permeated cities, villages and camps around Nepal on Saturday, \textcolor{blue}{after a massive 7.8 magnitude earthquake struck around midday.} \textcolor{ForestGreen}{Hours later, after a wave of relentless aftershocks, many people still were too scared to go back inside any buildings.} Others crowded around rubble, including men and women racing to rescue those trapped. And then there are the hundreds already confirmed dead, not to mention the hundreds more who suffered injuries. \textcolor{pink}{Below are some accounts from witnesses in the mountainous Asian nation, in their own words.} Fast Facts: Earthquakes . Anderson, an American who was in Nepal for trekking and meditation, was in his hotel room when the quake struck. "I went outside five minutes after the major tremors stopped. I went to a parking lot nearby for one hour or so, then walked down the main road," he said. He took a series of photos on the main road between Thamal and Durbar Squares, that he shared via CNN iReport. Kumar posted a photo of people in his neighborhood sheltering in a makeshift tent after the quake. He sent updates via Twitter about what he was seeing in the Lalitpur District of Kathmandu. "It's getting dark, no power and no water supply in Lalitpur area, but people are helping each other with food and other items . "Almost everyone staying outside home...Hard time for small kids and older people . "People are very worried and are planning to stay out on the street overnight, but they lack sufficient food and water." \textcolor{red}{Joshi is a UNICEF communication officer who was on the ground at the time of the quake. "The shake was like nothing I have experienced in my 57 years. It was strong and it shook for a long time."} \textcolor{orange}{Old monuments and temples fell, Joshi wrote of his experience. There were fears that other buildings would collapse.} "When I went out in the evening, I saw many people preparing to camp out in the main open parade ground in the middle of the street. Relatives were crying in the main government hospital where the dead were being lined up in front of the hospital building. "My family is traumatised. We are 5 generations living under one roof -- from a 100 year old grandmother to my 16 month old granddaughter. Strong aftershocks are keeping most of us up!" "Some of the historical sites are completely devastated. "Most of the people -- a lot of the people -- are walking through the city. They're confused and scared. A lot of people are crying. "They're out with their pets and their families and a lot of locals are volunteering in rescue operations. "In several parts of Kathmandu, a lot of people seem trapped under the rubble. Locals are trying to rescue these people because they can still hear them." Are you in Nepal or have loved ones affected? Please share with us if you are in a safe place. "We are scared and waiting for the tremors to end. We are all sitting outside because there is more news of another quake. "There is no power and families are listening to the FM radio inside their cars. News of multiple building collapses. "I've seen many cracked walls and roads and buildings. "The Dharahara was packed with people a while ago. There are police everywhere trying to move rubble to make space on the roads for ambulances. Everyone is very scared. " "I see many cracked buildings and people are panicked and all running down to the streets. \textcolor{cyan}{"The main landmark in Kathmandu is a spire, Dharahara, and it has fallen down, it is about 140 feet high in the center city.} "Another aftershock is hitting now, it is really strong. "Airplanes are circling now overhead and helicopters are flying and not clear if the airport is open.  We hear it is damaged." How are earthquakes measured? \textcolor{cyan}{"Many historic buildings have collapsed in the city. "In all my years I have never seen such a big earthquake here.} [...] \\ \hline
\textbf{GOLD SUMMARY}: \textcolor{blue}{Massive 7.8 magnitude earthquake has struck Nepal near its capital, Kathmandu .} As the death toll rises, witnesses describe devastation and panic . \\ \hline
\textbf{PEGASUS}: \textcolor{ForestGreen}{After hours of relentless aftershocks, many people still were too scared to go back inside any buildings.}  \textcolor{pink}{Below are some accounts from witnesses in the mountainous Asian nation, in their own words.} \textcolor{orange}{Old monuments and temples fell, and there were fears that other buildings would collapse.} \\ \hline
\textbf{BART}: Hundreds of people are dead and injured in Nepal on Saturday. \textcolor{ForestGreen}{Aftershocks and aftershocks have been felt in Kathmandu, Nepal's capital city.} \textcolor{red}{"The shake was like nothing I have experienced in my 57 years. It was strong and it shook for a long time"} \\ \hline
\textbf{Our model}: \textcolor{red}{'"The shake was like nothing I have experienced in my 57 years," a UNICEF worker says.} \textcolor{blue}{The 7.8-magnitude earthquake struck around midday Saturday.} \textcolor{ForestGreen}{Many people were too scared to go back inside any buildings.} \textcolor{cyan}{Many buildings in Kathmandu collapsed, including a 140-foot spire.'} \\ \hline
\end{tabular}
\end{small}
\caption{Another generated summary example}
\label{tab:example_2}
\end{table}}

\newpage
\section{Loss of Flow-based Neural Topic Model}
\label{app:loss_proof}

We have the loss of neural topic model, called evidence lower bound 

\begin{equation}
    \mathcal{L}_{\text{NTM}} = \log p(\mathbf{x,z}) - \log q(\mathbf{z}|\mathbf{x})
    \label{eq:ntm_loss}
\end{equation}

Let $f_1, f_2, …, f_K$ be a chain of invertible transformations which have Jacobian easy-to-compute. We have change of variable formula for transforming $\mathbf{z}_i$ to $\mathbf{z}_{i+1}$

\begin{equation}
    q(\mathbf{z}_{i+1}) = q(\mathbf{z}_i) \left| \det \frac{\partial f_{i+1}}{\partial \mathbf{z}_{i}} \right|^{-1}
\end{equation}

Sequentially applying for $K$ transformations, we have

\begin{equation}
    q(\mathbf{z}_{K}) = q(\mathbf{z}_{0}) \prod _{i=1}^{K} \left| \det \frac{\partial f_{i}}{\partial \mathbf{z}_{i-1}} \right|^{-1}
\end{equation}

or equivalently,

\begin{equation}
    \log q(\mathbf{z}_{K}) = \log q(\mathbf{z}_{0}) - \sum _{i=1}^{K} \left| \det \frac{\partial f_{i}}{\partial \mathbf{z}_{i-1}} \right|
    \label{eq:flow}
\end{equation}

Plugging the formula (\ref{eq:flow}) to equation (\ref{eq:ntm_loss}), we obtain

\begin{equation}
\begin{split}
    & \mathcal{L}_{\text{NTM}} = \log p(\mathbf{x,z}) - \log q(\mathbf{z}|\mathbf{x}) = \log p(\mathbf{x},\mathbf{z}_K) - \log q(\mathbf{z}_K|\mathbf{x}) \\
    &=  -\log q(\mathbf{z}_0) + \sum_{i=1}^{K} \log \abs{\det \frac{\partial f_i}{\partial z_{i-1}}} +\log p(\mathbf{x}|\mathbf{z}_K) + \log p(\mathbf{z}_K)
\end{split}
\end{equation}

We reach the neural topic model component in our training objective.

\section{Implementation Details}
\label{app:impl_details}

\textbf{Flow-based Neural Topic Model} Following \cite{wang2020friendly}, we preprocess to remove stopwords in the BoW representations. We experiment with different number of topics in {50, 100, 150, 200}  and the number of invertible transformations in flow-based neural topic model (flow length) on CNN/DailyMail dataset. The results (in the format of R1/R2/RL scores) are shown in Table \ref{tab:flow_topic_experiment}. It can be seen that a simple posterior distribution in neural topic model is not sufficient to describe the large-scale dataset, while a highly complex one can negatively affect the performance slightly. Similarly, it is necessary to set an adequate number of topics. We proceed to use flow length of 4 and topic number of 100 for other datasets.  \\ 

\begin{table}[h]
\centering
\begin{normalsize}
\begin{tabular}{c|c|c|c}
\hline
\textbf{Topic Num./Flow length} & 1 & 4 & 16 \\ \hline
50 & 44.19/21.49/41.28 & 44.48/21.54/41.36 & 44.23/21.51/41.29 \\ \hline
100 & 44.30/21.78/41.37 & \textbf{44.52/21.95/41.39} & 44.40/21.87/41.38 \\ \hline
150 & 44.25/21.70/41.34 & 44.44/21.86/41.27 & 44.30/21.80/41.21 \\ \hline
200 & 44.24/21.61/41.23 & 44.35/21.75/41.22 & 44.24/21.69/41.20 \\ \hline
\end{tabular}
\end{normalsize}
\caption{Comparisons on the number of topics and flow length on CNN/DailyMail dataset}
\label{tab:flow_topic_experiment}
\end{table}
	
We pretrain our versions of flow-based neural topic model on five downstream datasets CNN/DailyMail, XSum, Reddit TIFU, arXiv, and PubMed with batch size $\{256, 256, 256, 320, 320\}$, respectively. All versions are trained with Adadelta optimizer with a learning rate of $0.01$. \\

\noindent \textbf{Topic-oriented Transformer-based summarization model}. We do not change any settings from original papers of PEGASUS \cite{zhang2020pegasus} and BART \cite{lewis2019bart}. In particular, we finetune all models on 16 Nvidia GeForce A100 GPUs with batch size $256$, Adam optimizer of learning rate $1e-4$. For the objective function in Equation \ref{eq:loss}, we experimented $\lambda$ with the choices of $\{0.5, 0.6, 0.75, 0.9\}$ and found that $\lambda = 0.75$ gives the best performance for all datasets. Models are evaluated and saved checkpoints every one epoch. During training, we keep track three best validated checkpoints in terms of evaluation loss on the validation set. Eventually, for decoding, we run beam search with beam size of $8$ and note the best result out of three validated checkpoints. 

\end{document}